%% file: main.tex
\documentclass{article}

\usepackage{microtype}
\usepackage{graphicx}
\usepackage{subfigure}
\usepackage{booktabs} % for professional tables
\usepackage{booktabs}
\usepackage{makecell}
\usepackage{longtable}
\usepackage{multirow}
\usepackage{hyperref}

\usepackage[accepted]{icml2025}
\usepackage{amssymb}
\usepackage{mathtools}
\usepackage{amsthm}

\usepackage[capitalize,noabbrev]{cleveref}

\theoremstyle{plain}

\theoremstyle{definition}

\theoremstyle{remark}

\usepackage[disable,textsize=tiny]{todonotes}
\newcommand{\methodname}{\textsc{Sage Deer}}

\usepackage{colortbl}
\usepackage{enumitem}
\usepackage{multirow}
\usepackage{xcolor}

\definecolor{mygray}{RGB}{226, 226, 226}
\definecolor{myred}{RGB}{252, 142, 142}
\definecolor{mygreen}{RGB}{147, 255, 143}
\definecolor{myblue}{RGB}{144, 155, 255}
\definecolor{myyellow}{RGB}{253, 253, 143}
\definecolor{mypurple}{RGB}{255, 142, 250}

\icmltitlerunning{{\methodname}}

\begin{document}

\twocolumn[
\icmltitle{Sage Deer: A Super-Aligned Driving Generalist Is Your Copilot}



\icmlsetsymbol{equal}{*}

% Hao LU, Jiaqi Tang, Jiyao Wang, Yunfan LU, Xu Cao, Qingyong Hu, Yin Wang, Yuting Zhang, Tianxin Xie, Yunpeng Zhang, Yong Chen, Jiayu.Gao, Bin Huang, Dengbo He, Shuiguang Deng, Hao Chen, Ying-Cong Chen 

\begin{icmlauthorlist}
\icmlauthor{Hao LU}{equal,yyy}
\icmlauthor{Jiaqi Tang}{equal,yyy}
\icmlauthor{Jiyao Wang}{yyy}
\icmlauthor{Yunfan LU}{yyy}
\icmlauthor{Xu Cao}{yyy}
\icmlauthor{Qingyong Hu}{yyy}
\icmlauthor{Yin Wang}{sch}
\icmlauthor{Yuting Zhang}{yyy}
\icmlauthor{Tianxin Xie}{yyy}
\icmlauthor{Yunpeng Zhang}{phigent}
\icmlauthor{Yong Chen}{giga}
\icmlauthor{Jiayu Gao}{giga}
\icmlauthor{Bin Huang}{yyy}
\icmlauthor{Dengbo He}{yyy}
\icmlauthor{Shuiguang Deng}{sch}
\icmlauthor{Hao Chen}{yyy}
\icmlauthor{Ying-Cong Chen}{yyy}
\end{icmlauthorlist}

\icmlaffiliation{yyy}{HKUST, HKUST-GZ}
\icmlaffiliation{sch}{ZJU}
\icmlaffiliation{phigent}{Phigent}
\icmlaffiliation{giga}{GEELY. * means equal contribution}

\icmlcorrespondingauthor{Ying-Cong Chen}{yingcongchen@ust.hk}

\icmlkeywords{Machine Learning, ICML}

\vskip 0.3in
]



\printAffiliationsAndNotice  % leave blank if no need to mention equal contribution
% \printAffiliationsAndNotice{\icmlEqualContribution} % otherwise $use the standard text.

\begin{abstract}
\input{Sections/0_Abstract}
\end{abstract}

\input{Sections/1_Introduction}

\input{Sections/2_Preliminaries}

\input{Sections/3_Methodology}

\input{Sections/4_Experiments}

\input{Sections/6_Related_Work}

\input{Sections/7_Conclusion}

% \section*{Acknowledgements}

\bibliography{example_paper}
\bibliographystyle{icml2025}

%%%%%%%%%%%%%%%%%%%%%%%%%%%%%%%%%%%%%%%%%%%%%%%%%%%%%%%%%%%%%%%%%%%%%%%%%%%%%%%
%%%%%%%%%%%%%%%%%%%%%%%%%%%%%%%%%%%%%%%%%%%%%%%%%%%%%%%%%%%%%%%%%%%%%%%%%%%%%%%
% APPENDIX
%%%%%%%%%%%%%%%%%%%%%%%%%%%%%%%%%%%%%%%%%%%%%%%%%%%%%%%%%%%%%%%%%%%%%%%%%%%%%%%
%%%%%%%%%%%%%%%%%%%%%%%%%%%%%%%%%%%%%%%%%%%%%%%%%%%%%%%%%%%%%%%%%%%%%%%%%%%%%%%

%\input{Sections/Appendix}

%%%%%%%%%%%%%%%%%%%%%%%%%%%%%%%%%%%%%%%%%%%%%%%%%%%%%%%%%%%%%%%%%%%%%%%%%%%%%%%
%%%%%%%%%%%%%%%%%%%%%%%%%%%%%%%%%%%%%%%%%%%%%%%%%%%%%%%%%%%%%%%%%%%%%%%%%%%%%%%

\end{document}

%% file: Sections/0_Abstract.tex
 The intelligent driving cockpit, an important part of intelligent driving, needs to match different users' comfort, interaction, and safety needs. This paper aims to build a \textbf{\textcolor{gray}{s}}uper-\textbf{\textcolor{gray}{a}}ligned and \textbf{\textcolor{gray}{ge}}neralist \textbf{\textcolor{gray}{dr}}iving agent, \textbf{\textcolor{gray}{sage deer}}. Sage Deer achieves three highlights: (1) Super alignment: It achieves different reactions according to different people's preferences and biases. (2) Generalist: It can understand the multi-view and multi-mode inputs to reason the user's physiological indicators, facial emotions, hand movements, body movements, driving scenarios, and behavioral decisions. (3) Self-Eliciting: It can elicit implicit thought chains in the language space to further increase generalist and super-aligned abilities. Besides, we collected multiple data sets and built a large-scale benchmark. This benchmark measures the deer's perceptual decision-making ability and the super alignment's accuracy.

%% file: Sections/1_Introduction.tex
\section{Introduction}
% Background
The concept of a driving copilot refers to the collaboration between human drivers and vehicles to ensure seamless and efficient operations throughout the driving process~\cite{gptdriver,Cui2023DriveAY,Wang2020ASO,Hecht2018ARO}. As intelligent vehicles continue to evolve, this concept has become a cornerstone of their development, aiming to enhance safety, improve comfort, increase traffic efficiency, and deliver a superior driving experience for both drivers and passengers~\cite{Cui2023PersonalizedAD,Yang_2024_WACV}. For instance, driver monitoring is essential to proactively assess a driver's health, fatigue, mood, and behavior. Human-machine interaction plays a pivotal role in understanding driver actions, enabling the system to activate various cockpit service functions~\cite{li2023intelligent,yang2022comprehensive}. Furthermore, driving assistance features, such as lane-keeping and adaptive cruise control, help automate driving tasks, making the driving experience safer and more convenient.

\begin{figure}[!t]
\centering
\includegraphics[width=1\linewidth]{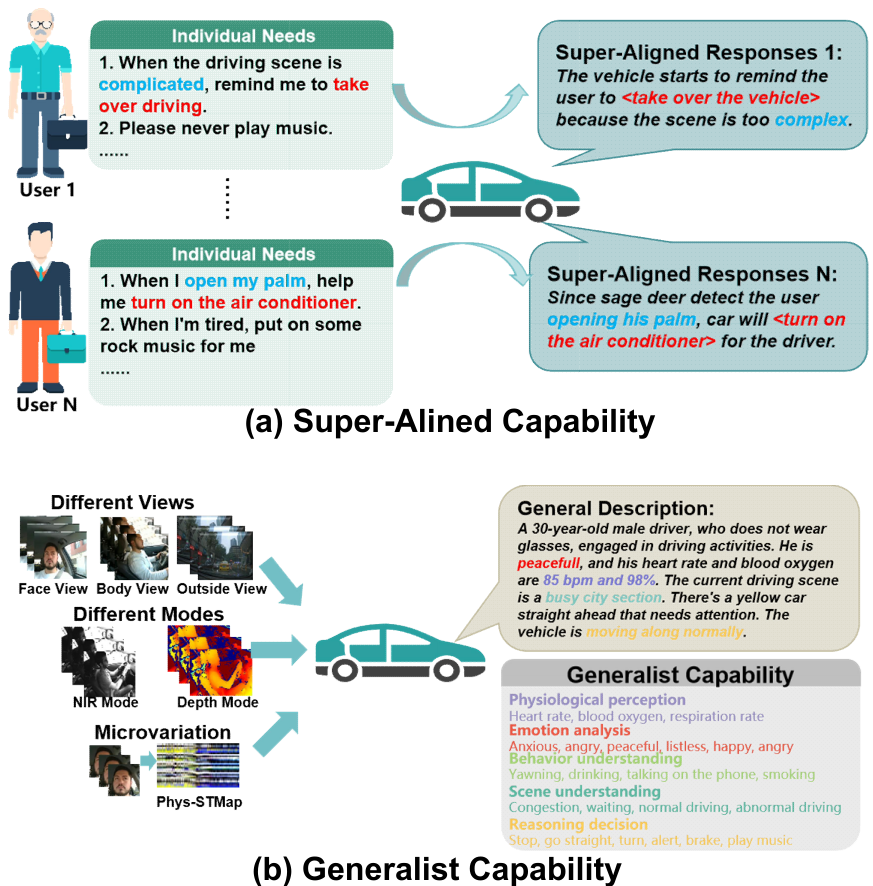}
\vspace{-7mm}
\caption{The Capability of Sage Deer. (a) Sage Deer can perform corresponding actions according to the preferences of each user. Users can save their preferences in a document and update them anytime. (b) Sage Deer can integrate multi-view and multi-model inputs for driver and scene understanding. Notably, Sage Deer actually has both super-aligned and generalist capabilities at the same time.} 
\label{fig:intro}
\end{figure}

% Super-Aligned
Although research on driving copilots has made progress, much of the prior work has focused on data collection, multi-task learning, and improving generalization performance. However, developing a super-aligned copilot model that addresses individual-specific requirements remains an underexplored area. \textbf{Super-aligned}, as defined in this paper, refers to the ability of a model to adapt to individual user requirements and perform corresponding actions, as illustrated in Fig.~\ref{fig:intro}(a). In addition, driving copilots must serve as generalists, capable of perception, understanding, and decision-making across diverse scenarios, as shown in Fig.~\ref{fig:intro}(b). In essence, Sage Deer seeks to understand multi-view and multi-modal inputs while adapting its behavior to meet user-specific needs.

Multimodal large language models (MLLMs) give a great opportunity to show strong capabilities in integrating multimodal inputs, handling multitask outputs, and demonstrating generalization abilities. However, most of these approaches are not well-suited for driving copilot applications, with limitations in data, benchmarking, and model design. Specifically, (1) MLLMs must feed the user's needs and habits into the large model to achieve the super-aligned ability through long text at every inference. (2) Most of the MLLMs can not be satisfied and can accept multi-view and multi-mode input at the same time. (3) Super-aligned and generalist abilities tend to overfit rather than reason through visual instruction tuning. (4) Satisfying both of these requires large-scale tailored training data of the driving copilot, which is also lacking.

To address these limitations, we propose a novel \textbf{\textcolor{gray}{s}}uper-\textbf{\textcolor{gray}{a}}ligned and \textbf{\textcolor{gray}{ge}}neralist \textbf{\textcolor{gray}{dr}}iving copilot framework, called \textbf{\textcolor{gray}{Sage Deer}}. For achieving generalist capabilities, we introduce multi-view tokenizers, multimodal tokenizers, and microvariation tokenizers. For super-aligned capabilities, we design a learnable retrieval-augmented generation (RAG) framework capable of delivering personalized responses based on user-specific needs. Additionally, we propose a Continuous Latent Chain Elicitation (CLCE) mechanism to further enhance both generalist and super-aligned capabilities. The CLCE activates the inherent reasoning capabilities of LLMs by stimulating implicit chains-of-thought (COT) reasoning without requiring explicit COT labels. Lastly, we compile and combine multiple datasets to construct a comprehensive benchmark for evaluating driving copilots. In summary, our key contributions are as follows:

1. We introduce a novel super-aligned generalist driving copilot framework capable of understanding multi-view and multi-modal inputs while providing personalized responses tailored to individual user needs.

2. We design a Continuous Latent Chain Elicitation mechanism to strengthen the copilot’s super-aligned and generalist capabilities by leveraging the LLM’s innate reasoning abilities without relying on additional COT annotations.

3. We establish a multi-view and multi-modal evaluation protocol for driving copilots that unifies the assessment of driver physiology, emotion, behavior, scene understanding, and decision-making.

%% file: Sections/2_Preliminaries.tex
\section{Data Curating and Beachmarking}

\subsection{Data collection}

Our goal is to create a super-aligned driving generalist in the intelligent copilot. We selected two of the most recent multi-view multi-task large-scale driving datasets AIDE~\cite{yang2023aide} and DMD~\cite{ortega2020dmd}.To contactless monitor the user's health condition, five remote physiological measurement datasets (VIPL-HR~\cite{niu2019robust}, V4V~\cite{revanur2021V4V}, PURE~\cite{PURE2014}, BUAA-rPPG~\cite{xi2020BUAA} and UBFC~\cite{UBFC2017}) is used. To quantitatively evaluate the fatigue state of drivers, YawDD~\cite{abtahi2014yawdd} dataset is used in this paper. Combined with these datasets, sage deer can satisfy multi-view and multi-modal input (RGB, NIR and Depth), generalist capabilities (physiological estimation, emotional estimation, gesture estimation, behavior estimation, driving behavior detection, and driving decision-making), and super-aligned capabilities.

\begin{figure*}[!t]
\centering
\includegraphics[width=1\linewidth]{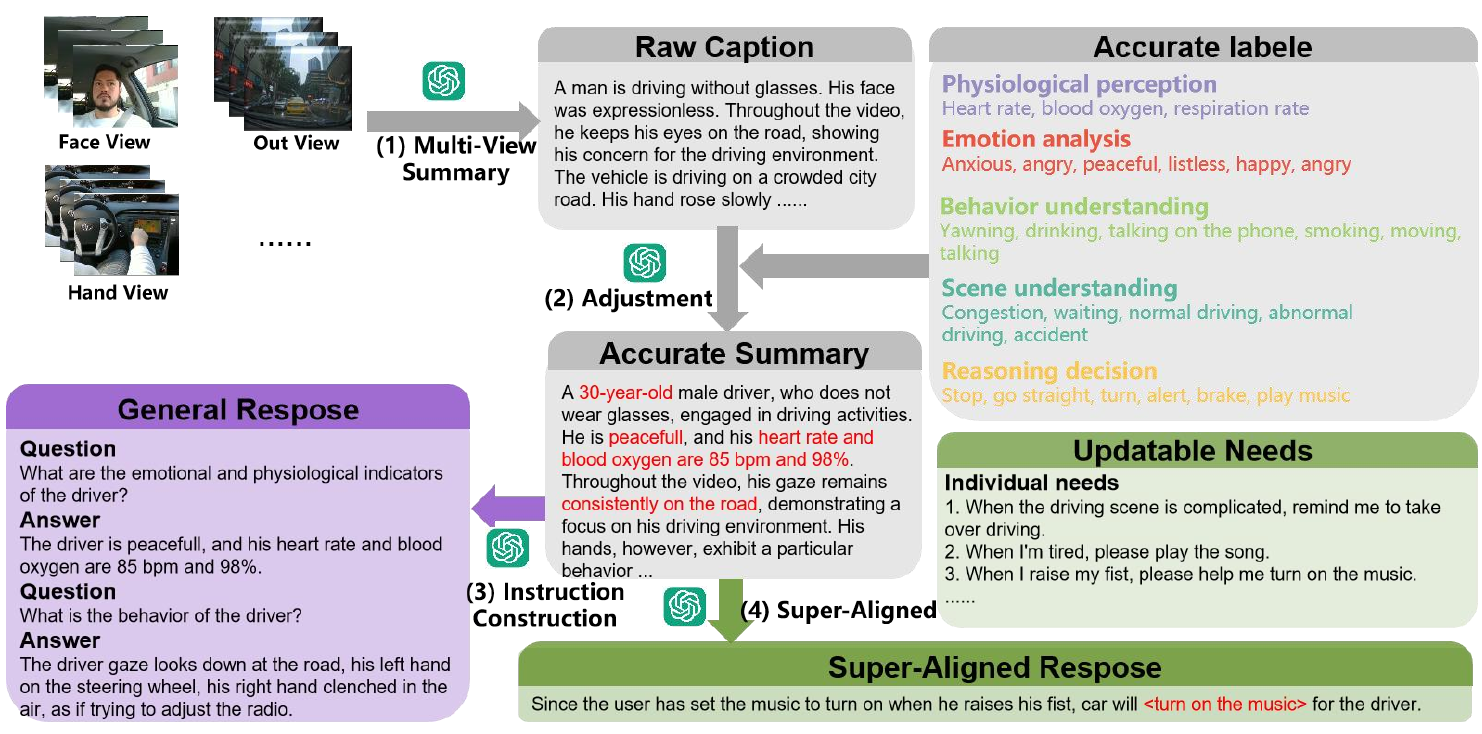}
\vspace{-6mm}
\caption{Data construction process of Sage Deer. (1) We use existent GPT-4o tools to generate captions for the videos likes~\citep{li2023videochat,damonlpsg2023videollama}. Then, we set a reasonable prompt to merge the information from different videos. (2) We took advantage of the existing labels (including physiological indicators, emotional indicators, action indicators, behavioral indicators, scene understanding, and reasoning decision-making) to correct and supplement the captain. (3) Next, we use GPT4~\citep{achiam2023gpt} as an assistant to build question-answering pairs for different tasks (including physiological indicators, emotion, behavior, and so on.).  (4) We design multiple user preferences, and GPT4 responds to the current scenario based on user preferences. All of the above processes are double-checked manually.} 
\label{Data}
\end{figure*}

\subsection{General Instruction Construction}

To construct natural language descriptions for sage deer, we built a data curating pipeline. Firstly, we use existing GPT-4o tools to generate captions for the frame sampled at equal intervals automatically. Likes~\citep{li2023videochat,damonlpsg2023videollama}, we set the reasonable prompt to merge the information of different frames. Existing image caption methods are often inaccurate or insufficiently annotated for an intelligent copilot system. So, we took advantage of the existing tags (including physiological indicators, emotional indicators, action indicators, behavioral indicators, scene understanding, and reasoning decision-making) in the dataset to correct and supplement the captain. Next, we use GPT-4o as an assistant to build question-answering pairs for different tasks.

% See supplementary materials~\ref{data}.

\subsection{Super-Aligned Reaction}

Users have different requirements for the driving copilot, especially interactivity and trustworthiness. Based on interactivity and trustworthiness, we simulate and design a variety of different requirements levels to build requirements documents for each user. In addition, we have built different replies in interactivity and trustworthiness. Interactivity means creating gestures, emotions, and body movements and tailoring copilot feedback to user needs. Trustworthiness refers to personalized warning feedback on user fatigue, bad mood, and bad behavior. Based on different requirements documents, the same scene GPT-4o will be used to generate different interactivity and trustworthiness responses. 

%Please refer to the supplementary material~\ref{data} for details and examples.

%% file: Sections/3_Methodology.tex
\section{Framework}
\label{headings}

\begin{figure*}[h]
\begin{center}
\includegraphics[width=0.96\linewidth]{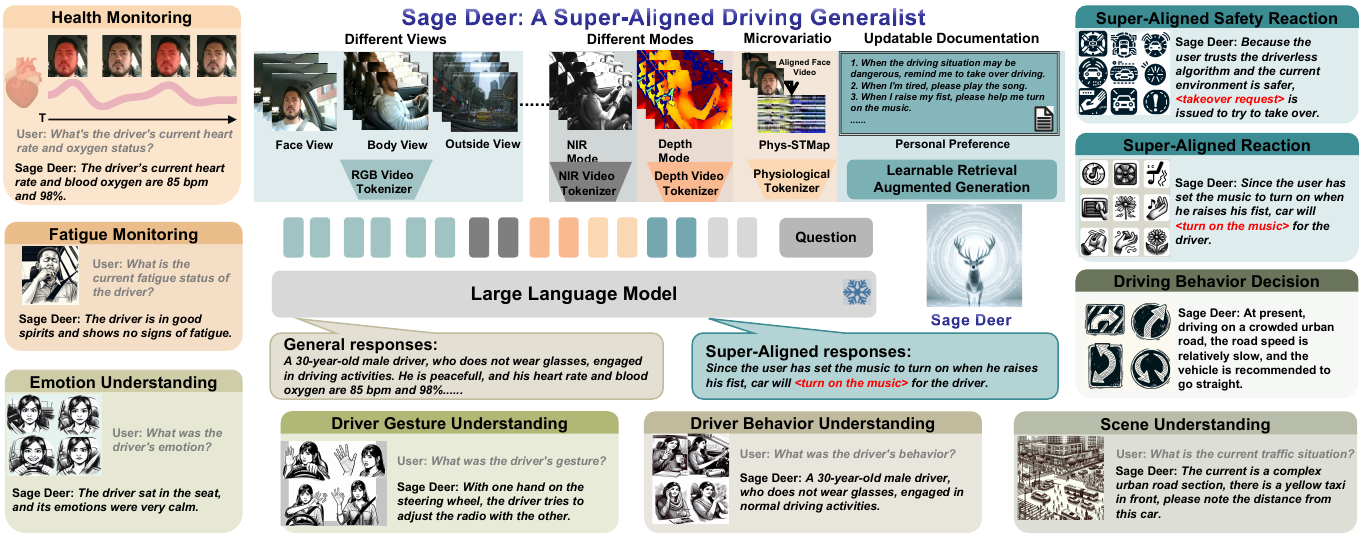}
\end{center}
%\vspace{-5mm}
\caption{
Sage Deer uses pre-trained video encoders (visual tokenizers) to tokenize the different modes and views of the video, especially the physiological encoder used to extract the physiological signals of the face video. Using tokenized visual embedding, large language models can provide general responses for physiological indicators, emotional states, gestures, human behavior, and scene understanding. It is worth emphasizing that the individual needs of the user can be edited in a single document. We then query this information with learnable retrieval augmentation generation, generating results super aligned with user preferences.
}
%\vspace{-6mm}
\label{intro}
\end{figure*}

We aim to build a super-aligned generalist in the driving copilot as shown in Fig.~\ref{intro}. Specifically, generalist ability is the ability to integrate multi-view and multi-mode information and detect the driver's physiological indicators, emotions, behaviors, and scene understanding. Super-aligned gives responses based on the user's preferences. We highlight that Sage Deer's framework is simple and necessary without a complex design.

\subsection{The Generalist Capability}
\label{general}
Leveraging multi-view and multi-mode input for driving copilot can significantly enhance the model's ability to understand complex scenes, especially under challenging conditions such as low-light environments.  

\textbf{Tokenizing Multi-View.} Multi-view input is crucial for understanding both the driving scene and the driver against occlusion. Like most MLLMs~\cite{gao2023llama,lin2023video,sima2023drivelm,li2023videochat,wang2022internvideo}, we utilized the CLIP as the feature extractor from multi-view images.  Using a two-layer linear network, the extracted features were then mapped to the language space features $em \in \mathbb{R}^{C \times L}$, where the $L$ is the number of language tokens occupied by visual features. Additionally, we employed markers to distinguish between different views. For example, a front-facing RGB video embedding is denoted $em_{front} \in \mathbb{R}^{C \times L}$, formatted as $E_{front} = < \text{Front RGB bos} > \, em_{front} \, < \text{Front RGB cos} >$, here $< \text{Front RGB bos} >$ and $< \text{Front RGB cos} >$ are actually a tokenizered vectors. This format ensures that the LLM can understand exactly which view the feature is coming from.

\textbf{Tokenizing Multi-Model.} For the driving copilot, understanding and reasoning in adverse conditions require a combination of multiple sensing modes, including near-infrared (NIR) and depth images. We uniformly employed the CLIP as the feature extractor for both NIR and depth. Since CLIP is specifically designed for RGB images, we set up the first layer of CLIP as a fine-tuned retraining process. Subsequently, a simple two-layer linear network mapped these mode features into the language space features $em \in \mathbb{R}^{C \times L}$. Here $C$ represents the number of channels characteristic of the language model, and $L$ is a hyperparameter indicating the number of tokens used to represent the video features. To understand different modes, we also incorporate makers. For instance, we appended specific start and end symbols to the NIR video embeddings, formatted as $E+{NIR} = < \text{NIR bos} > \, em_{NIR} \, < \text{NIR cos} >$. Here $< \text{NIR bos} >$ and $< \text{NIR cos} >$ are actually tokenized vectors. Similar processing was applied to other video modalities before feeding them into the LLM.

\textbf{Tokenizing Microvariation.} In addition to the ability to observe drivers' expressions, emotions, and behaviors, Sage Deer also needs to measure further the driver's health status, such as heart rate (HR), breathing, and heart rate variability (HRV). With remote photoplethysmography (rPPG)  technology, we can obtain these indicators through the non-contact RGB video~\cite{wang2015novel,xiao2024remote,lu2023neuron,niu2019robust,liu2024rppg}. Here, we select NEST-rPPG as our physiological tokenizer, which is trained on multiple datasets~\cite{lu2023neuron}. Fellow NEST-rPPG, we pre-trained a neural network that could convert video into HR, breathing, and HRV, and then we kept only the encoder as our physiological tokenizer. Subsequently, a simple two-layer linear network mapped these mode features into the language space features $em_{phys} \in \mathbb{R}^{C \times L}$. Similarly, we set up a special makers $E_{phys} = < \text{Physiological bos} > \, em_{phy} \, < \text{Physiological cos} >$.

Along with questions and responses, multi-view, multi-mode, and microvariation tokenization can be jointly fed into LLM, i.e. $\{E_{front}, E_{out}, E_{face}, E_{hand}, E_{NIR}, E_{Depth}, ...., E_{phys},\\ E_{rag}, <bos>, Q, E_{cot}, R, <cos>\}$. $Q$ and $R$ are the tokenized questions and responses. $E_{rag}$ is the super-aligned feature based on the current scene and the user's personal preferences. $E_{cot}$ helps the LLM understand and fuse these multiple inputs through implicit COT, which is described in Sec.~\ref{CLCE}. There's only one phase of training, and the main loss is visual instruction tuning $\mathcal{L}_{vi}$~\citep{li2023videochat,damonlpsg2023videollama,Maaz2023VideoChatGPT}.

\subsection{Super-Aligned Capability}

Super-aligned refers to the model's ability to align with individual requirements to carry out corresponding actions. However, we cannot fine-tune every MLLM model for each user's different requirements. To this end, we have adopted a simple retrieval-augmented generation (RAG) framework as shown in Fig.~\ref{intro}.

Specifically, we built an up-to-date document that includes personal requirements. Then, we chunked the document in a very concise way to divide it into $N$ sentences. In other words, a chunk is a sentence. Each sentence is then tokenized and then filled with empty to the same length $M$, i.e., $em_{sa} \in \mathbb{R}^{N \times M \times C}$, where $C$ is the length of the vocabulary table. A four-layer convolutional encoder is then used to compress the length of sentence tokens. Then, we calculate the similarity between visual features and sentence features and the weighted combination of these sentence tokens $E_{rag} \in \mathbb{R}^{\times K \times C}$. $em_{rag}$ reflects the individual requirements, which is also fed into the LLM with other multi-modal information as explained at the end of Sec.~\ref{general}.

\section{Continuous Latent Chain Eliciting}
\label{CLCE}

\begin{figure*}[!t]
\centering
\includegraphics[width=1\linewidth]{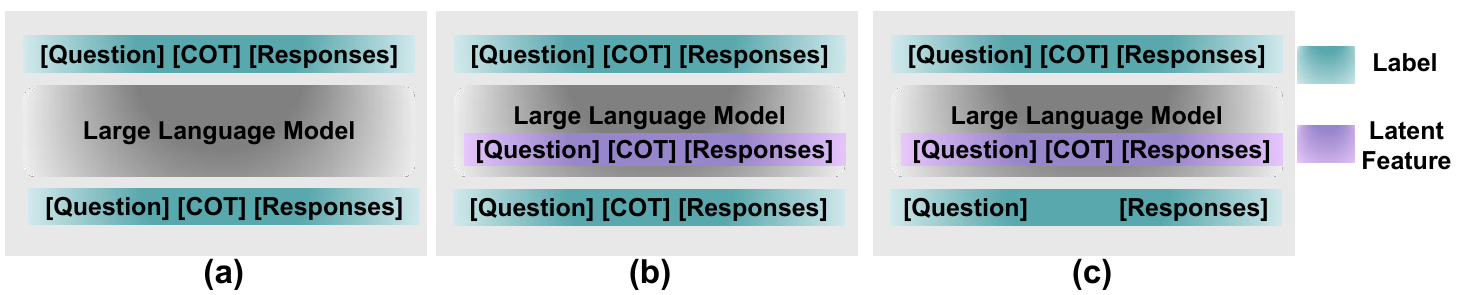}
\vspace{-7mm}
\caption{A comparison of our continuous latent chain eliciting with the existent chain-of-thought method. (a) The traditional CoT model generates the reasoning process with the supervision of COT tokens. (b) Coconut uses the LLM to reason in an unrestricted latent space instead of a language space~\cite{hao2024training}. But he still distills knowledge from the COT label. (c) Our method tries to activate the implicit language space so that the model learns the implicit COT from itself.} 
\label{fig:CLCE}
\end{figure*}

To achieve super-aligned generalists, Sage Deer must simultaneously fuse multi-view and multi-modal information, align individual preferences, and model the task relationship. A reasonable chain of thought (COT) can efficiently make these processes efficient in reasoning in the language space. However, the accuracy label of the COT process for cockpit tasks is missing. Inspired by~\cite{hao2024training}, we propose the Continuous Latent Chain Eliciting (CLCE) strategy, as shown in Fig.~\ref{fig:CLCE}.

 \begin{figure}[!ht]
\centering
\includegraphics[width=1\linewidth]{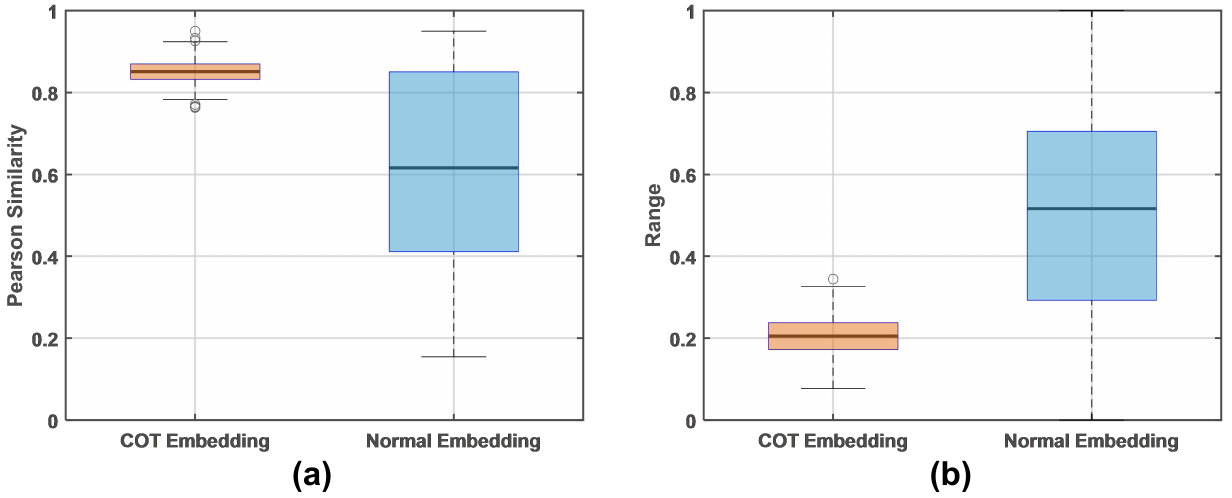}
\vspace{-7mm}
\caption{The trivial solutions of the CL-COT embedding $E_{cot}$. (a) Compared with normal embedding, COT embedding tends to be self-similar. (b) Compared with normal embedding, COT embedding tends to have a smaller range of values. These two phenomena indicate that the COT embedding tends to be invalid.} 
\label{fig:cot}
\end{figure}

\textbf{Continuous Latent Chain-of-Thought (CL-COT).} The CLCE forces the LLM to output the fixed $L$ length CL-COT before the final responses. The fixed CL-COT is expected to learn implicit reasoning without any label supervision. The principle behind this method is that it forces the LLM to think of more COT steps in the hidden space to get the final answer. However, CL-COT input is required because the LLM's equal length of input and output can speed up training. So, we feed multi-modal, multi-view, and rag features $\{E_{front}, E_{out}, ..., E_{rag}\}$ into a simple 2-layer convolution network to output a learnable CL-COT embedding $E_{cot}$. $E_{cot}$ is a self-learning COT embedded in the LLM hidden space, which is expected to activate the LLM's self-reasoning ability.

\textbf{Latent Chain Eliciting.} The learnable COT embedding $E_{cot}$ exhibits trivial solutions in our experiment as shown in Fig.~\ref{fig:cot}. Specifically, we visualize the Pearson similarity and range of random two token features of Continuous Latent Chain (CLC) embedding. Compared with normal embedding, the similarity between CLC tokens is very high, and the difference between the maximum and minimum values of the same token is very small. This shows that the CLC token $E_{cot}$ tends to predict an invalid token rather than the valid COT process. Here we use a very simple method to solve this problem. We design the latent chain eliciting (LCE) loss $\mathcal{L}_{LCE}$ to make the constraint token valid, i.e., $\mathcal{L}_{LCE} = - \| E_{cot} \|_1$. LCE loss forces the LLM to output an active embedding, and the final response labels ensure that this active embedding is a valid COT process.

%% file: Sections/4_Experiments.tex
\section{Experiments}
\label{others}

\subsection{Datasets.} We found two recent data sets of very comprehensive annotations for assisted driving. The DMD is a driver monitoring dataset, an extensive dataset that includes real and simulated driving scenarios: distraction, eye distribution, drowsiness, handwheel interaction, and contextual data, in 41 hours of RGB, depth, and infrared video from 3 cameras, capturing the faces, bodies, and hands of 37 drivers~\cite{ortega2020dmd}. AIDE proposes an assisted driving awareness dataset that takes into account contextual information both inside and outside the vehicle in a natural scenario~\cite{yang2023aide}. AIDE enables overall driver monitoring through multi-perspective Settings of the driver and the scene, multi-modal annotations of the face, body, posture, and gestures, and four practical task designs for driver understanding. we collect five rPPG face video datasets ( VIPL-HR~\cite{niu2019rhythmnet}, PURE~\cite{stricker2014non}, UBFC-rPPG~\cite{bobbia2019unsupervised}, V4V~\cite{revanur2021first}, and BUAA-MIHR~\cite{xi2020image}), mostly with subjects remaining still, and some with head movements.

\subsection{Training Details.} We tain our model on A6000 for 2 epochs. The learning rate and weight decay are set to 0.001 and 0.02, respectively. The maximum sentence length is set to 64. That is, if the sentence is too long, excess parts will be discarded, and if the sentence is too short, 0 will be filled to make the length uniform. For the feature extraction of physiological signals, we use the pre-trained model of NEST-rPPG~\cite{lu2023neuron} and do not fine-tune it.

\subsection{Baselines and Evaluation Metrics}
We compare our proposed framework's understanding performance with SOTA video understanding baselines. We select five baselines: Video-ChatGPT~\citep{Maaz2023VideoChatGPT}, VideoChat~\citep{li2023videochat}, Video-LLaMA~\citep{damonlpsg2023videollama}, LLaMA-Adapter~\citep{zhang2023llamaadapter}, and Video-LLaVA~\citep{lin2023video}. These methods do not have a multi-perspective, multi-modal understanding, so we migrated our generalist tokenizer method to them and fine-tuned it. The main purpose of comparing these methods is to emphasize the capabilities of our CLCE.

To evaluate our model's performance accurately, we adopt BLEU Bilingual Evaluation Understudy(BLEU) and SPICE~\citep{papineni2002bleu} to measure word overlap between the model-generated text and the ground truth. The consistently higher scores across both metrics validate the effectiveness of our approach in generating coherent, accurate, and contextually rich descriptions.

\begin{table*}[ht]
    \centering
    \caption{Generalist capabilities on AIDE data sets.}
    \label{tab:metrics_per_category}
    \begin{tabular}{lcccccccc}
        \toprule
        \multirow{2}{*}{\textbf{Method}} & \multicolumn{2}{c}{\textbf{Emotion}} & \multicolumn{2}{c}{\textbf{Behavior}} & \multicolumn{2}{c}{\textbf{Scene}} & \multicolumn{2}{c}{\textbf{Condition}} \\
        \cmidrule(lr){2-3} \cmidrule(lr){4-5} \cmidrule(lr){6-7} \cmidrule(lr){8-9}
         & BLEU & SPICE & BLEU & SPICE & BLEU & SPICE & BLEU & SPICE \\
        \midrule
        \textbf{Video-ChatGPT} & 0.200 & 0.280 & 0.170 & 0.205 & 0.190 & 0.330 & 0.195 & 0.320 \\
        \textbf{VideoChat} & 0.205 & 0.285 & 0.175 & 0.210 & 0.195 & 0.335 & 0.200 & 0.325 \\
        \textbf{Video-Llama} & 0.217 & 0.312 & 0.183 & 0.223 & 0.207 & 0.352 & 0.211 & 0.337 \\
        \textbf{Llama-Adapter} & 0.215 & 0.310 & 0.190 & 0.225 & 0.210 & 0.340 & 0.189 & 0.321 \\
        \textbf{Video-LLaVA} & 0.202 & 0.290 & 0.172 & 0.205 & 0.190 & 0.330 & 0.195 & 0.320 \\
        \textbf{Ours} & \textbf{0.232} & \textbf{0.331} & \textbf{0.194} & \textbf{0.242} & \textbf{0.225} & \textbf{0.369} & \textbf{0.223} & \textbf{0.360} \\
        \bottomrule
    \end{tabular}
    \label{AIDE}
\end{table*}

\begin{table*}[ht]
    \centering
    \caption{Generalist capabilities on DMD data sets.}
    \label{tab:metrics_per_category}
    \begin{tabular}{lcccccc}
        \toprule
        \multirow{2}{*}{\textbf{Method}} & \multicolumn{2}{c}{\textbf{Action}} & \multicolumn{2}{c}{\textbf{Gaze}} & \multicolumn{2}{c}{\textbf{Hand}} \\
        \cmidrule(lr){2-3} \cmidrule(lr){4-5} \cmidrule(lr){6-7}
         & BLEU & SPICE & BLEU & SPICE & BLEU & SPICE  \\
        \midrule
        \textbf{Video-ChatGPT} & 0.175 & 0.250 & 0.150 & 0.185 & 0.160 & 0.310  \\
        \textbf{VideoChat} & 0.190 & 0.260 & 0.160 & 0.185 & 0.160 & 0.310 \\
        \textbf{Video-Llama} & 0.205 & 0.275 & 0.170 & 0.205 & 0.180 & 0.325 \\
        \textbf{Llama-Adapter} & 0.202 & 0.265 & 0.170 & 0.201 & 0.175 & 0.320 \\
        \textbf{Video-LLaVA} & 0.195 & 0.295 & 0.155 & 0.220 & 0.170 & 0.315 \\
        \textbf{Ours} & \textbf{0.235} & \textbf{0.315} & \textbf{0.195} & \textbf{0.230} & \textbf{0.210} & \textbf{0.340} \\
        \bottomrule
    \end{tabular}
    \label{DMD}
\end{table*}
\subsection{Generalist Performance}

Our model can estimate the driver's emotion, physiological indicators, gaze, physical behavior, hand behavior, driving scene, and vehicle state. In order to more clearly evaluate the ability of the model in different subtasks, we conducted a systematic evaluation on two multi-task datasets, fatigue and physiological indicators. The AIDE dataset evaluates generalist capabilities across four categories—emotion, behavior, scene, and condition. As shown in Table~\ref{AIDE}, our model consistently outperforms all baselines, achieving higher BLEU and SPICE scores across all categories. This improvement indicates the model's enhanced ability to capture nuanced details such as emotional states and contextual anomalies. The integration of expert knowledge through RAG (Retrieval-Augmented Generation) serves as a critical factor in bridging connections between related tasks, leading to superior results.

\begin{figure}[!t]
\centering
\includegraphics[width=1\linewidth]{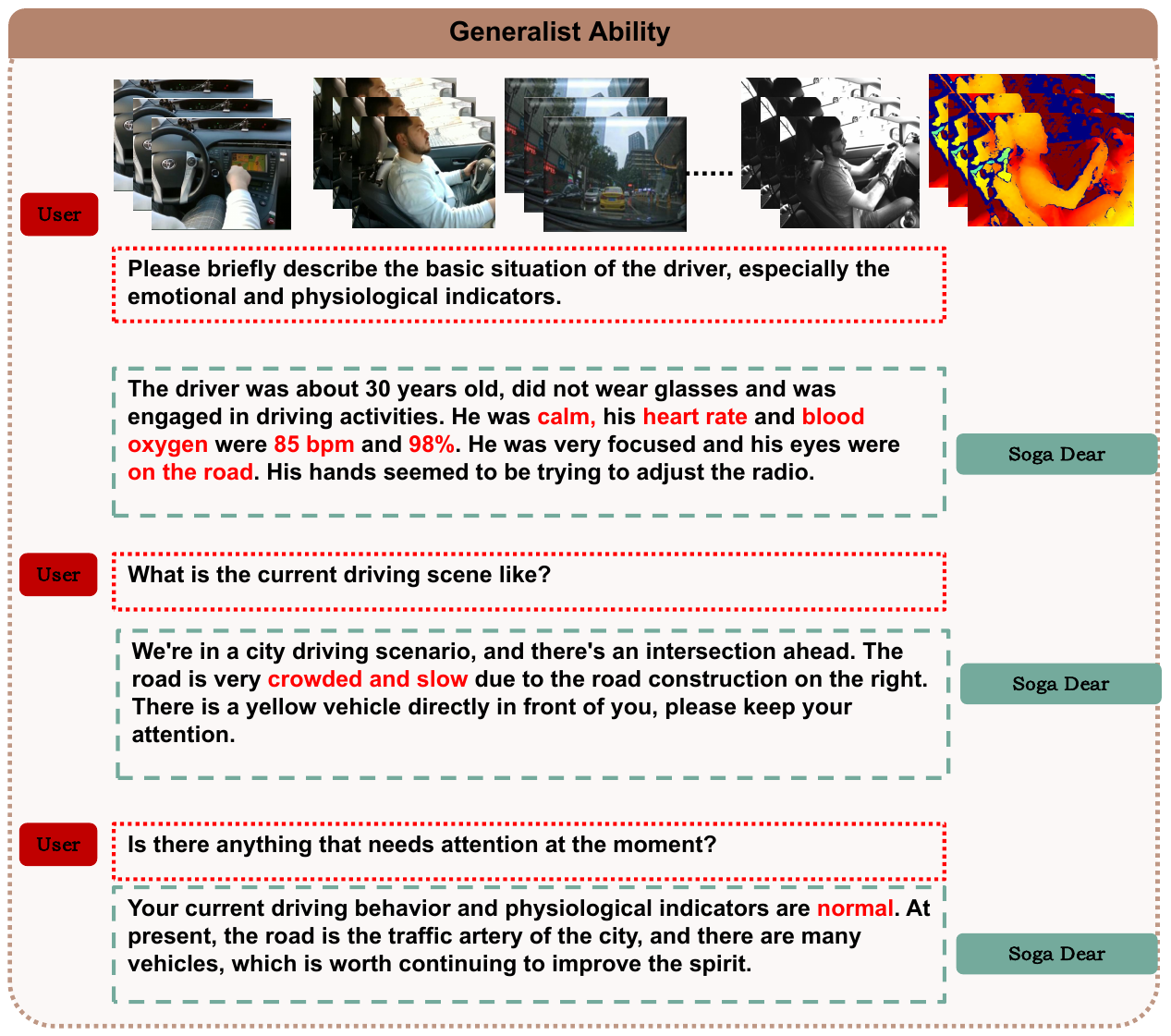}
\vspace{-6mm}
\caption{Generalist ability. Soga deer can reply to different tasks and can reply to an open domain.} 
\vspace{-6mm}
\label{ga}
\end{figure}

\begin{figure}[!t]
\centering
\includegraphics[width=1\linewidth]{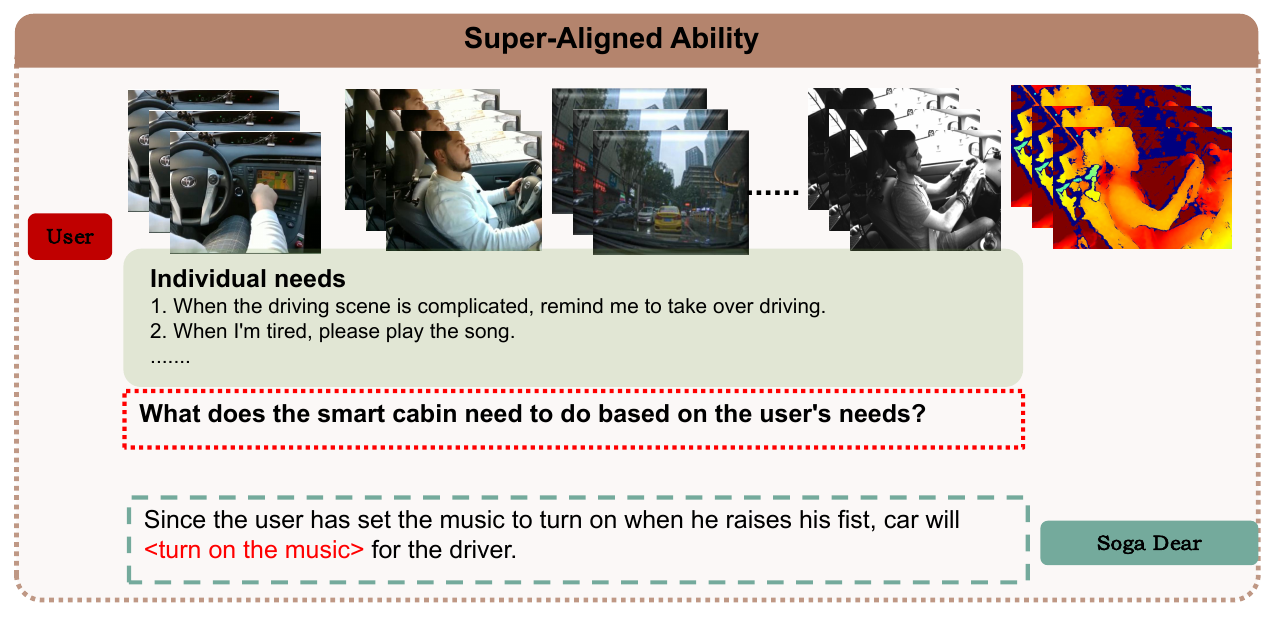}
\vspace{-6mm}
\caption{Super alignment ability. soga deer can perform different operations according to the individual needs of each person.} 
\label{sa}
\end{figure}

Further, we verify our algorithm on the DMD data set. The DMD dataset focuses on three categories: action, gaze, and hand behavior. As shown in Table~\ref{DMD}, our model demonstrates remarkable performance improvements over the baselines, further showcasing its generalist capabilities. Particularly, the model performs exceptionally well in understanding and describing complex interactions involving gaze and hand movement. This suggests that our framework excels at parsing fine-grained details critical for tasks involving driver monitoring and safety analysis.

\subsection{Super-Aligned Performance}
\label{saaaa}

For driving cockpits, users have diverse and individualized preferences, necessitating personalized super-alignment capabilities. However, the majority of existing algorithms lack support for this level of alignment. A straightforward baseline approach involves encoding all user preferences as text inputs to a large language model (LLM) and leveraging the LLM's long-range reasoning capabilities. In contrast, our method achieves super-alignment through a learned Retrieval-Augmented Generation (RAG) framework. The comparative performance of these two approaches is presented in Table~\ref{sa}.

\begin{table}[t!]
    \centering
    \caption{Baseline compared the performance of our approach on the super alignment protocol. The baseline is to user requirements directly into visual features and feed them into the LLM, taking advantage of the long-range modeling capabilities of the LLM itself.}
    \label{tab:metrics_per_category}
    \begin{tabular}{lcccc}
        \toprule
        \multirow{2}{*}{\textbf{Method}} & \multicolumn{2}{c}{\textbf{AIDE}} & \multicolumn{2}{c}{\textbf{DMD}} \\
        \cmidrule(lr){2-3} \cmidrule(lr){4-5} 
         & BLEU & SPICE & BLEU & SPICE \\
        \midrule
        \textbf{Baseline} & 0.184 & 0.301 & 0.191 & 0.296\\
        \textbf{Ours} & \textbf{0.231} & \textbf{0.327} & \textbf{0.215} & \textbf{0.314}\\
        \bottomrule
    \end{tabular}
    \label{sa}
\end{table}

As shown in Table~\ref{sa}, our approach achieves significant improvements over the baseline. These results highlight the limitations of LLMs in handling long text inputs and making nuanced responses solely based on visual features and prior knowledge. This underscores the necessity of our RAG framework for achieving robust super-alignment.

\subsection{Ablation Study}
\begin{figure}[!t]
\centering
\includegraphics[width=1\linewidth]{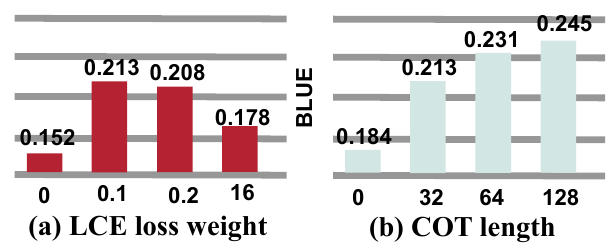}
\vspace{-6mm}
\caption{Ablation Study. (a) The impact of  LCE loss weight. (b) The impact of COT length. } 
\label{fig:ablation}
\end{figure}

Sage Deer's core innovation is the CLCE approach. Therefore, we mainly conduct ablation experiments on the hyperparameters of this method. Here we compare the weight of LCE loss and the compression of implicit COT length for ablation as shown in Fig.~\ref{fig:ablation}. The experimental results show that a reasonable weight of LCE loss can enable LLM to stimulate COT ability in the form of hidden language space. The longer the length of the implicit COT, the better the model performance.

%% file: Sections/6_Related_Work.tex
\section{Related Work}
\label{gen_inst}

\begin{table*}[t]
\label{tab}
\centering
\caption{Comparison of public driving cockpit datasets. Reaction is the response given by the intelligent cockpit according to different user needs. Question Answering refers to the verbal inquiry interaction between the user and the cockpit.}    
\resizebox{\linewidth}{!}{ 
\begin{tabular}{ccccccccccc}
\toprule
Dataset    & Views & Multimodal & Health & Behavior & Emotion & Traffic Context & Vehicle Condition & Reaction & Question Answering \\ \midrule

SEU~\cite{zhao2012recognition}        & 1     & \textbf{--} & \textbf{--} & $\checkmark$ & \textbf{--} & \textbf{--} & \textbf{--} & \textbf{--} & \textbf{--} \\

Tran  \cite{tran2018real}  & 1    & \textbf{--} & \textbf{--} & $\checkmark$ & \textbf{--} & \textbf{--} & \textbf{--} & \textbf{--} & \textbf{--} \\

Zhang  \cite{zhang2020driver} & 2    & $\checkmark$ & \textbf{--} & $\checkmark$ & \textbf{--} & \textbf{--} & \textbf{--} & \textbf{--} & \textbf{--} \\

StateFarm \cite{statefram}  & 1    & \textbf{--} & \textbf{--} & $\checkmark$ & \textbf{--} & \textbf{--} & \textbf{--} & \textbf{--} & \textbf{--} \\

AUC-DD~\cite{eraqi2019driver}     & 1   & \textbf{--} & \textbf{--} & $\checkmark$ & \textbf{--} & \textbf{--} & \textbf{--} & \textbf{--} & \textbf{--} \\ 

LoLi~\cite{saad2020end}       & 1    & $\checkmark$ & \textbf{--} & $\checkmark$ & \textbf{--} & \textbf{--} & \textbf{--} & \textbf{--} & \textbf{--} \\

Brain4Cars~\cite{jain2016brain4cars}       & 2    & \textbf{--} & \textbf{--} & $\checkmark$ & \textbf{--} & \textbf{--} & \textbf{--} & \textbf{--} & \textbf{--} \\

Drive\&Act~\cite{martin2019drive} & 6    & $\checkmark$ & \textbf{--} & $\checkmark$ & \textbf{--} & \textbf{--} & \textbf{--} & \textbf{--} & \textbf{--} \\

DMD~\cite{ortega2020dmd}        & 3     & $\checkmark$ & \textbf{--} & $\checkmark$ & \textbf{--} & \textbf{--} & \textbf{--} & \textbf{--} & \textbf{--} \\

DAD~\cite{kopuklu2021driver}       & 2     & $\checkmark$ & \textbf{--} & $\checkmark$ & \textbf{--} & \textbf{--} & \textbf{--} & \textbf{--} & \textbf{--} \\

DriPE~\cite{guesdon2021dripe}      & 1   & \textbf{--} & \textbf{--} & \textbf{--} & \textbf{--} & \textbf{--} & \textbf{--} & \textbf{--} & \textbf{--} \\

LBW~\cite{kasahara2022look}       & 2   & \textbf{--} & \textbf{--} & \textbf{--} & \textbf{--} & \textbf{--} & \textbf{--} & \textbf{--} & \textbf{--} \\

MDAD~\cite{jegham2019mdad}       & 2    & $\checkmark$ & \textbf{--} & $\checkmark$ & \textbf{--} & \textbf{--} & \textbf{--} & \textbf{--} & \textbf{--} \\

3MDAD~\cite{jegham2020novel}      & 2   & $\checkmark$ & \textbf{--} & $\checkmark$ & \textbf{--} & \textbf{--} & \textbf{--} & \textbf{--} & \textbf{--} \\

DEFE~\cite{li2021spontaneous}       & 1    & \textbf{--} & \textbf{--} & \textbf{--} & $\checkmark$ & \textbf{--} & \textbf{--} & \textbf{--} & \textbf{--} \\

DEFE+~\cite{li2021cogemonet}     & 1    & $\checkmark$ & \textbf{--} & \textbf{--} & $\checkmark$ & \textbf{--} & \textbf{--} & \textbf{--} & \textbf{--} \\

Du  ~\cite{du2020convolution}    & 1    & $\checkmark$ & \textbf{--} & \textbf{--} & $\checkmark$ & \textbf{--} & \textbf{--} & \textbf{--} & \textbf{--} \\

KMU-FED~\cite{jeong2018driver}    & 1    & \textbf{--} & \textbf{--} & \textbf{--} & $\checkmark$ & \textbf{--} & \textbf{--} & \textbf{--} & \textbf{--} \\

MDCS~\cite{oh2022multimodal}       & 2   & $\checkmark$ & \textbf{--} & \textbf{--} & $\checkmark$ & \textbf{--} & \textbf{--} & \textbf{--} & \textbf{--} \\ 

\textbf{AIDE~\cite{yang2023aide}}      & 4   & \textbf{--} & \textbf{--} & $\checkmark$ & $\checkmark$ & $\checkmark$ & $\checkmark$ & \textbf{--} & \textbf{--} \\ 

\midrule 

\textbf{Sage Deer}      & 4   & $\checkmark$ & $\checkmark$ & $\checkmark$ & $\checkmark$ & $\checkmark$ & $\checkmark$ & $\checkmark$ & $\checkmark$ \\ 
\bottomrule
\end{tabular}
}
\end{table*}
\paragraph{Multimodal Large Language Model}

Recent research highlights the robust capabilities of large models in video understanding, showcasing significant progress in this area. Beyond the advancements in vision-language large models~\cite{li2023blip,zhu2023minigpt,liu2024visual,lu2024gpt}, there is a growing emphasis on integrating additional modalities, such as audio and sensor data, to further enhance model performance~\cite{lv2024video,li2023videochat,maaz2023video,ye2023mplug,luo2023valley}. For instance, Bain et al.~\cite{Bain21} introduced a large-scale dataset offering general video content descriptions, pushing the boundaries of how models can interpret dynamic visual scenes. Several LLM-based approaches~\cite{li2023videochat,maaz2023video,ye2023mplug,luo2023valley} aim to effectively interpret the visual elements in videos, unlocking deeper contextual understanding. Furthermore, Video-LLaMa~\cite{damonlpsg2023videollama} expands the scope of video comprehension by incorporating both visual and auditory modalities, enhancing the richness of information these models can process. Additionally, Su et al.~\cite{su2023pandagpt} have leveraged multimodal encoders that work across six distinct modalities, pushing the envelope on the versatility of multimodal models. However, despite these exciting advancements, there remains a notable gap in research focused on multimodal large models tailored specifically for driving cockpits.

\paragraph{Multi-modality Large Model in Driving}

Multi-modality large models have been widely applied in autonomous driving systems, supporting tasks such as automatic planning and control~\cite{gptdriver,Cui2023DriveAY}, perception~\cite{Wang2020ASO}, and driver health monitoring~\cite{Hecht2018ARO}, among others. By integrating multi-modal data (e.g., vision, speech, point clouds, etc.)~\cite{Yang2023LLM4DriveAS}, these models enhance the ability of autonomous driving systems to perceive both the internal and external environments. In user-facing vehicle interfaces, analyzing driver-specific factors such as emotions, health, posture, and actions allows autonomous systems to interact dynamically and adapt to diverse driving styles. Cui et al.~\cite{Cui2023ReceiveRA,Cui2023DriveAY} were the first to propose leveraging multi-modal data from both the vehicle's interior and exterior to enhance decision-making precision in autonomous driving. Recent advancements~\cite{Cui2023PersonalizedAD,Yang_2024_WACV} have sought to improve fine-grained comprehension of user-specific language inputs and optimize vehicle control performance. Nonetheless, these approaches overlook the influence of latent factors such as emotional state, health conditions, and physical actions on vehicle safety and operational efficiency.

% In this paper, we super-align various modal information of users, which not only includes the user's own interaction, but also perceives the user's health, emotions, posture, behavior, and other information, thereby enhancing the intelligence of the vehicle system.

\paragraph{Retrieval-Augmented Generation}

Retrieval-Augmented Generation (RAG) enhances the accuracy of knowledge-intensive tasks by retrieving relevant information from an external knowledge base that can be continuously updated~\cite{Gao2023RetrievalAugmentedGF}. Specifically, RAG begins by collecting external information, archiving previous user queries~\cite{Ma2023QueryRF}, and organizing the knowledge into a structured repository provided by the user. An efficient indexing mechanism is then applied to retrieve image and textual data. Once relevant information is retrieved based on user-provided prompts—through an optimized combination of priority ranking and retrieval algorithms—RAG integrates the query with the selected knowledge, processes it through a large model, and generates a more accurate response. This capability enables RAG to address user-specific queries while simultaneously maintaining a coherent dialogue by aggregating historical interaction contexts. For this paper, we deploy a learnable RAG framework to store and utilize customized user data, including individual driving habits and behaviors. This facilitates the development of a more human-centered vehicle driving agent that can adapt to the unique requirements of specific users.

%% file: Sections/7_Conclusion.tex
\section{Conclusion}

This paper presents the development of a novel driving copilot framework, Sage Deer, designed to provide a super-aligned and generalist solution for intelligent vehicles. The framework integrates multi-view and multi-modal inputs, adapting to individual user preferences and needs while maintaining strong perception, understanding, and decision-making capabilities across diverse driving scenarios. Additionally, a Continuous Latent Chain Elicitation (CLCE) mechanism is proposed to enhance both super-aligned and generalist abilities by tapping into the inherent reasoning capabilities of large language models (LLMs).

% including driver physiology, emotion, behavior, scene understanding, and decision-making. This work contributes to advancing the field of driving assistants by offering a flexible, adaptive solution capable of improving safety, comfort, and overall driving experience.

% Sage Deer addresses key challenges in current driving copilot research, particularly the need for personalized responses and the integration of multimodal inputs. To achieve super-aligned capabilities, the authors introduce a learnable retrieval-augmented generation (RAG) framework, which tailors responses based on user-specific data. Additionally, a Continuous Latent Chain Elicitation (CLCE) mechanism is proposed to enhance both super-aligned and generalist abilities by tapping into the inherent reasoning capabilities of large language models (LLMs). The paper also discusses the creation of a comprehensive evaluation benchmark, combining multiple datasets to assess various aspects of driving copilot performance, including driver physiology, emotion, behavior, scene understanding, and decision-making. This work contributes to advancing the field of driving assistants by offering a flexible, adaptive solution capable of improving safety, comfort, and overall driving experience.


% \section*{Software and Data}